\PassOptionsToPackage{table}{xcolor}
\documentclass[10pt,twocolumn,letterpaper]{article}

\usepackage[pagenumbers]{cvpr}      

\usepackage{booktabs}
\usepackage{multirow}
\usepackage{graphicx}
\usepackage{tabularx}
\usepackage{array}
\usepackage{arydshln}
\usepackage{pifont}
\usepackage{adjustbox}
\usepackage[table]{xcolor}

\definecolor{checkgreen}{RGB}{46,125,90}
\definecolor{crossred}{RGB}{180,70,70}
\definecolor{lightgray}{RGB}{245,245,245}
\definecolor{myblue}{RGB}{214,228,240}

\newcommand{\cmark}{\textcolor{checkgreen}{\ding{51}}}
\newcommand{\xmark}{\textcolor{crossred}{\ding{55}}}

\usepackage[table]{xcolor}
\usepackage{booktabs}
\usepackage{colortbl}
\definecolor{linegray}{RGB}{185,195,205}
\definecolor{groupblue}{RGB}{235,242,248}
\definecolor{oursblue}{RGB}{225,238,248}
\definecolor{checkgreen}{RGB}{45,110,80}
\definecolor{crossgray}{RGB}{145,145,145}

\usepackage[most]{tcolorbox}
\usepackage{xcolor}

\definecolor{findingteal}{HTML}{00635F}
\definecolor{findingblue}{HTML}{557D9A}

\newtcolorbox{findingbox}[1][findingteal]{
    enhanced,
    colback=gray!5,
    colframe=#1,
    boxrule=0.6pt,
    arc=2mm,
    left=8pt,
    right=8pt,
    top=4pt,
    bottom=4pt,
    before skip=5pt,
    after skip=6pt,
    fontupper=\itshape,
    shadow={1.5mm}{-1.5mm}{0mm}{black!12}
}

\definecolor{cvprblue}{rgb}{0.21,0.49,0.74}
\usepackage[pagebackref,breaklinks,colorlinks,allcolors=cvprblue]{hyperref}

\def\confName{CVPR}
\def\confYear{2027}

\title{OmniV-Bench: Can MLLMs Comprehend Immersive Spatio-temporal Scenes?}

\author{
Liu Yang$^{1}$,
Huiyu Duan$^{1}$,
Yixuan Gao$^{1}$,
Sirui Liu$^{1}$,
Lu Liu$^{1}$,
Zhaolin Cai$^{1}$,
Ke Gu$^{2}$,\\
Xiongkuo Min$^{1}$,
Xiaoping Zhang$^{3}$,
Guangtao Zhai$^{1}$\\[3pt]
$^{1}$Shanghai Jiao Tong University, Shanghai, China\\
$^{2}$Beijing University of Technology, Beijing, China\\
$^{3}$Tsinghua University, Beijing, China\\[3pt]
{\ttfamily\small
\{ylyl.yl, huiyuduan, gaoyixuan, sirui\_liu\_116,
lettieliu, minxiongkuo,}\\
{\ttfamily\small
zhaiguangtao\}@sjtu.edu.cn,
107552301311@stu.xju.edu.cn, guke@bjut.edu.cn, xpzhang@ieee.org}
}

\begin{document}
\twocolumn[{
\vspace{-1em}
\maketitle
\centering
\vspace{-1em}
\includegraphics[width=1\linewidth]{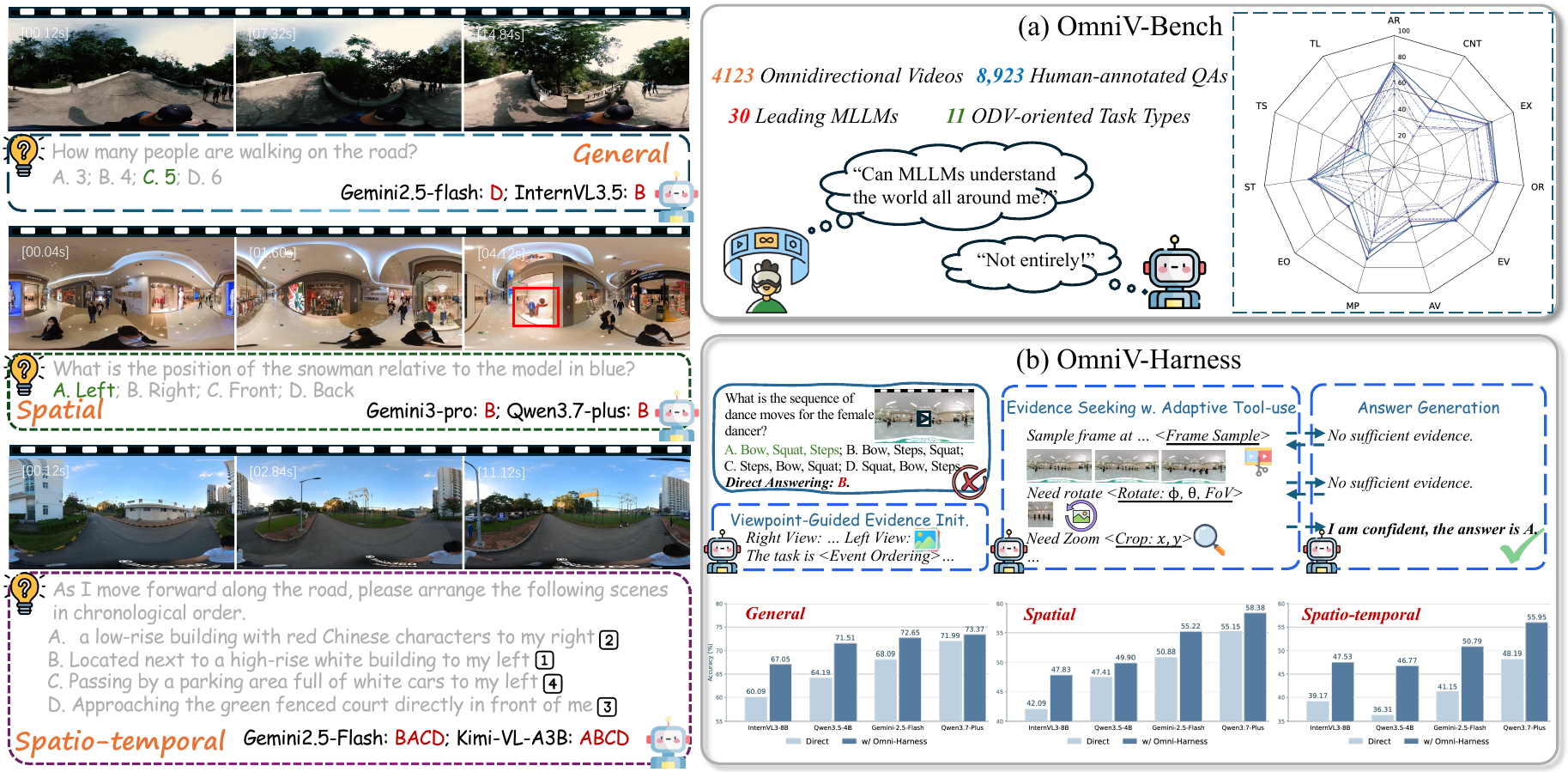}
\vspace{-2em}
\captionof{figure}{(a) We introduce OmniV-Bench, the first comprehensive
benchmark for evaluating MLLMs' capabilities in understanding
immersive omnidirectional videos covering general,
spatial, and spatio-temporal dimensions.
Extensive evaluations of leading MLLMs reveal substantial limitations in omnidirectional video understanding. Representative examples on the left
illustrate the challenges across these three
task categories.
(b) To enhance ODV understanding ability, we further propose OmniV-Harness, a multimodal harness that performs viewpoint-guided evidence initialization and task-relevant evidence acquisition with agentic visual tool use,
consistently improving MLLMs' performance across all three
task categories.
}
\label{fig:1_frontpage}
\vspace{1.5em}
}]

\begin{table*}[t]
\setlength{\belowcaptionskip}{-0.01cm}
\centering
\belowrulesep=0pt
\aboverulesep=0pt
\renewcommand{\arraystretch}{1.05}
\vspace{-1.5em}
\caption{
Comparison between widely adopted spatial understanding benchmarks,
omnidirectional image benchmarks, omnidirectional video benchmarks,
and our OmniV-Bench.
}
\vspace{-3mm}
\label{tab1:comparison}

\resizebox{0.95\linewidth}{!}{
\footnotesize
\begin{tabular}{l cccc ccc c}
\toprule[1.1pt]
\multirow{2}{*}{\textbf{Benchmark}}
& \multirow{2}{*}{\textbf{\#Visuals}}
& \multirow{2}{*}{\textbf{\#QA / Tasks}}
& \multirow{2}{*}{\textbf{\#Task Types}}
& \multirow{2}{*}{\textbf{Input Modality}}
& \multicolumn{3}{c}{\textbf{Dimension}}
& \multirow{2}{*}{\textbf{QA Source}} \\
\cmidrule(lr){6-8}
& & & & & \textbf{General} & \textbf{Spatial}
& \textbf{Spatio-temporal} & \\

\arrayrulecolor{linegray}
\specialrule{0.4pt}{1.5pt}{1.5pt}
\arrayrulecolor{black}
\rowcolor{groupblue}
\textit{\textbf{General Spatial Understanding Benchmarks}}
& & & & & & & & \\
ViewSpatial-Bench~\cite{li2026viewspatial}
& 1,000 & 5,700 & 5 & 3D Scene & \xmark & \cmark & \xmark & Auto \\
VSI-Bench~\cite{yang2025thinking}
& 29 & 5,000 & 8 & NFOV Video & \xmark & \cmark & \cmark & Auto \\
SSRBench~\cite{liu2026ssr}
& 789 & 789 & 6 & 2D Image & \cmark & \cmark & \xmark & Auto \\

\arrayrulecolor{linegray}
\specialrule{0.4pt}{1.5pt}{1.5pt}
\arrayrulecolor{black}
\rowcolor{groupblue}
\textit{\textbf{Omnidirectional Image Benchmarks}}
& & & & & & & & \\
VQA 360$^{\circ}$~\cite{chou2020visual}
& 1,490 & 17,000 & 6 & Indoor ODI & \cmark & \cmark & \xmark & Auto \\
OSR-Bench~\cite{dongfang2026multimodal}
& 4,100 & 153,000 & 3 & Indoor ODI & \xmark & \cmark & \xmark & Auto \\
Dense360-Bench~\cite{zhou2025dense360}
& 1,279 & 6,000 & 2 & Indoor/Outdoor ODI & \cmark & \xmark & \xmark & Auto \\
ODI-Bench
& 2,000 & 4,254 & 10 & Indoor/Outdoor ODI & \cmark & \cmark & \xmark & Manual/Auto \\

\arrayrulecolor{linegray}
\specialrule{0.4pt}{1.5pt}{1.5pt}
\arrayrulecolor{black}
\rowcolor{groupblue}
\textit{\textbf{Omnidirectional Video Benchmarks}}
& & & & & & & & \\
Pano-AVQA~\cite{yun2021pano}
& 5,400 & 51,700 & 2 & ODV + Audio & \xmark & \cmark & \xmark & Manual/Auto \\
360CityArena~\cite{watanabe2026360cityarena}
& 602 & 175 & 7 & Outdoor ODV & \cmark & \cmark & \xmark & Manual \\
\rowcolor{oursblue}
\textbf{OmniV-Bench (Ours)}
& \textbf{4,123} & \textbf{8,923} & \textbf{11}
& \textbf{Indoor/Outdoor ODV}
& \cmark & \cmark & \cmark & \textbf{Manual} \\
\bottomrule[1.1pt]
\end{tabular}
}
\vspace{-1.5em}
\end{table*}
\begin{abstract}
Omnidirectional videos (ODVs) capture dynamic environments with a full $360^\circ$ field of view,
providing rich spatial context and temporal dynamics
for applications such as virtual reality (VR),
augmented reality (AR), and embodied intelligence. Recent multimodal large language models (MLLMs) have made great progress in spatial and temporal understanding, yet their ability to comprehend the immersive 360$^{\circ}$ spatio-temporal information conveyed by ODVs remains largely unexplored. To bridge this gap, we first introduce \textbf{OmniV-Bench}, a \underline{Bench}mark designed to comprehensively evaluate the capabilities of MLLMs on spatio-temporal \underline{Omni}directional \underline{V}ideo understanding. OmniV-Bench contains 4,123 high-quality ODVs collected from diverse scenes, along with 8,923 human-annotated question-answer pairs spanning 11 fine-grained tasks across three categories, i.e., general, spatial, and spatio-temporal understanding. Extensive benchmark experiments reveal that the spatio-temporal complexity of ODVs poses significant challenges for current MLLMs. To address these limitations, we further propose \textbf{OmniV-Harness}, an agentic harness that enhances ODV understanding through viewpoint-guided evidence initialization and task-relevant evidence acquisition with adaptive tool use. Comprehensive experiments demonstrate that OmniV-Harness consistently improves immersive spatio-temporal reasoning ability of MLLMs. Both OmniV-Bench and OmniV-Harness will be released upon publication.
\end{abstract}
\vspace{-2em}
    
\section{Introduction}
\label{sec:intro}

Omnidirectional videos (ODVs) capture dynamic environments across a full spherical 360$^{\circ}$ field of view (FoV). Compared with conventional 2D videos or multi-view images, ODVs enable smooth immersive observation of the surrounding scene over time. Thus, ODVs are widely used in virtual reality (VR), augmented reality (AR), spatial navigation, and hold great potential for embodied perception and scene understanding~\cite{yang2025omni2,zheng2026learning}.

Although recent advances in multimodal large language models (MLLMs) have led to significant progress in spatial and temporal understanding of conventional videos and multi-view images~\cite{yang2025thinking,yang2026mmsi,lin2025mmsi}, research on omnidirectional media understanding has predominantly focused on static images~\cite{yang2026odi,dongfang2026multimodal,zhou2025dense360}, leaving omnidirectional video understanding largely underexplored. Existing ODV understanding benchmarks~\cite{watanabe2026360cityarena,yun2021pano} are limited in either task scope or evaluation objective, as summarized in Table~\ref{tab1:comparison}, leaving comprehensive and multidimensional ODV
understanding insufficiently evaluated. Compared with static omnidirectional images, ODVs introduce rich temporal dynamics, while offering
more comprehensive spatial context within
180$^{\circ} \times$ 360$^{\circ}$ immersive environments
than conventional videos. ODV understanding therefore requires jointly perceiving spatial relationships in omnidirectional environments and reasoning about how these relationships evolve over time.
These unique spatio-temporal characteristics pose additional challenges beyond both omnidirectional image understanding and conventional video understanding, highlighting the need for systematically evaluating MLLMs under ODV settings. 

To address this gap, we first introduce \textbf{OmniV-Bench}, a novel \underline{Bench}mark designed to comprehensively evaluate the \underline{Omni}directional \underline{V}ideo understanding capabilities of MLLMs, as shown in Figure~\ref{fig:1_frontpage} (a). OmniV-Bench comprises 4,123 high-quality ODVs covering diverse indoor and outdoor scenes, together with 8,923 human-annotated QA pairs spanning 11 fine-grained tasks across three complementary dimensions, including general understanding, spatial reasoning, and spatio-temporal reasoning. This benchmark enables a systematic evaluation of immersive ODV understanding.
Based on OmniV-Bench, we conduct extensive evaluation among various general-purpose and spatial-centric MLLMs. Benchmark results reveal that current MLLMs face substantial challenges in understanding immersive omnidirectional environments, particularly in spatial and spatio-temporal reasoning tasks.

To further improve 360$^{\circ}$ spatio-temporal reasoning ability of MLLMs, we introduce \textbf{OmniV-Harness}, the first agentic harness tailored for omnidirectional video understanding.
OmniV-Harness first employs viewpoint-guided evidence initialization to provide direction-aware guidance for subsequent spatio-temporal reasoning. 
It then performs adaptive evidence acquisition, enabling MLLMs
to actively acquire additional evidence through agentic visual tool use. This design allows models to selectively explore task-relevant timestamps, viewing directions, and spatial regions, acquiring task-relevant spatio-temporal evidence to refine their predictions before producing the final response.
Experiments on OmniV-Bench demonstrate that OmniV-Harness consistently improves both proprietary and open-source MLLMs across different model scales. The main highlights of this work include:
\begin{itemize}
    \item We propose \textbf{OmniV-Bench}, a comprehensive benchmark for evaluating MLLMs on omnidirectional video understanding, comprising 4,123 high-quality ODVs and 8,923 human-annotated QA pairs across 11 fine-grained general, spatial and spatio-temporal tasks.
    \item We conduct extensive evaluations of general-purpose and spatial-centric MLLMs on ODV comprehension tasks, revealing their substantial limitations to comprehend immersive
    spatio-temporal environments.
    \item We propose \textbf{OmniV-Harness}, an agentic harness that performs viewpoint-guided
    evidence initialization and task-relevant evidence acquisition with adaptive visual tool use. 
    \item Extensive experiments demonstrate the effectiveness and generalization of OmniV-Harness on both proprietary and open-source models.
\end{itemize}
\vspace{-0.5em}
\section{Related Works}
\vspace{-0.5em}
\label{sec:related_works}
\begin{figure}[t]\centering
\includegraphics[width=1\linewidth]{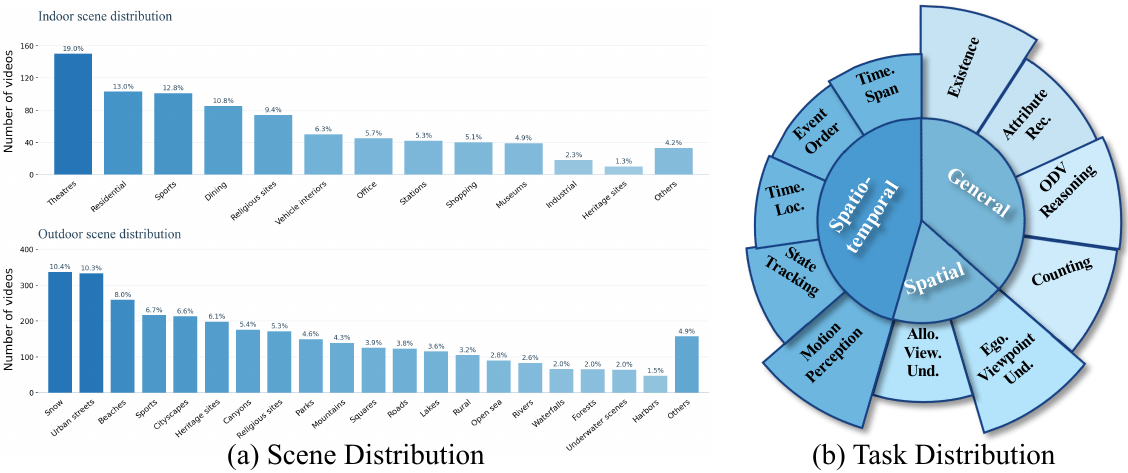}
\vspace{-1em}
\caption{Data distribution of OmniV-Bench.
}
\label{fig:3}
\vspace{-1em}
\end{figure}
\subsection{General Spatial Understanding Benchmarks}
With the rapid advancement of MLLMs, multimodal evaluation has expanded from general-level understanding~\cite{li2024mvbench,liu2024mmbench,wang2025lvbench} toward spatial-level understanding~\cite{li2026viewspatial,yang2025thinking,liu2026ssr}. Existing spatial understanding benchmarks typically evaluate MLLMs using perspective images, multi-view images, or normal-field-of-view (NFOV) videos, as summarized in Table~\ref{tab1:comparison}. However, such visual inputs typically capture only
partial observations of the surrounding environment. In contrast, omnidirectional images provide complete scene coverage within a single frame, while omnidirectional videos further extend such immersive observations over time.
The combination of holistic spatial context and temporal dynamics makes ODV understanding and its evaluation particularly valuable for a wide range of applications, including immersive VR/AR, spatial navigation, autonomous perception, embodied intelligence, and dynamic scene analysis. 

\subsection{Omnidirectional Image and Video Understanding Benchmarks}
Several benchmarks have been proposed for omnidirectional image and video understanding, 
as summarized in Table~\ref{tab1:comparison}.
Existing studies predominantly focus on 
ODI understanding. Dense360-Bench~\cite{zhou2025dense360} evaluates captioning and grounding, while ODI-Bench~\cite{yang2026odi} covers both general and spatial understanding across diverse scenes.
However, research on ODV understanding remains limited, with existing benchmarks focusing on specific tasks and scenarios. 
Pano-AVQA~\cite{yun2021pano} targets spherical spatial and audio-visual reasoning, whereas 360CityArena~\cite{watanabe2026360cityarena} focuses on embodied urban navigation in outdoor urban environments. As a result, comprehensive ODV understanding,
particularly the joint reasoning over immersive
spatial context and temporal dynamics, remains
insufficiently evaluated. Our OmniV-Bench addresses this gap by providing a comprehensive evaluation of general, spatial, and spatio-temporal understanding across diverse ODV scenes.
\section{OmniV-Bench} 
\subsection{Overview}
We introduce OmniV-Bench, a benchmark designed to comprehensively evaluate MLLMs on omnidirectional video understanding. OmniV-Bench comprises 4,123 high-quality indoor and outdoor ODVs, together with 8,923 human-annotated question-answer pairs across 11 fine-grained general, spatial and spatio-temporal tasks, enabling a systematic evaluation of MLLMs in complex, immersive, and dynamically evolving environments.

\subsection{Fine-grained Task Definition}
Existing ODV benchmarks primarily focus on specialized
settings, such as audio-visual question answering or embodied navigation~\cite{watanabe2026360cityarena,yun2021pano}, leaving the unique spatial and spatio-temporal characteristics of ODVs insufficiently explored. In contrast, OmniV-Bench aims to systematically assess the general-purpose ODV comprehension capabilities of MLLMs across three complementary dimensions, \textit{i.e.}, general, spatial and spatio-temporal, with the task distribution shown in Figure~\ref{fig:3}(b).
\vspace{-1em}
\paragraph{General-level ODV understanding}
We introduce four general-level ODV understanding tasks, \textit{i.e.}, \textit{attribute recognition, counting, existence, and ODV reasoning}, to evaluate the basic comprehension capabilities of MLLMs. These tasks do not explicitly require complex spatial or spatio-temporal reasoning. Instead, their primary challenge lies in perceiving dense visual information distributed across omnidirectional scenes. \textit{Attribute recognition} evaluates fine-grained perception by requiring models to localize task-relevant objects in complex scenes and identify their properties, such as color, material, and appearance. \textit{Counting} and \textit{existence} assess global perception by requiring models to comprehensively inspect the omnidirectional scene to determine object quantities or verify the presence of specific objects. \textit{ODV reasoning} further evaluates the ability to integrate visual evidence with semantic and real-world knowledge to infer plausible scene-level conclusions.
\vspace{-2em}
\paragraph{Spatial-level ODV understanding}
This dimension composes of two spatial-level tasks, \textit{i.e.}, \textit{egocentric viewpoint understanding} and \textit{allocentric viewpoint understanding}, These tasks do not necessarily require temporal reasoning but emphasize spatial localization and
viewpoint-dependent reasoning. \textit{Egocentric viewpoint understanding} evaluates spatial relations with respect to the VR viewer or ego camera, while \textit{allocentric viewpoint understanding}
requires models to reason from the perspective of
another entity in the scene.
\vspace{-1.5em}
\paragraph{Spatio-temporal-level ODV understanding}
Beyond general perception and spatial reasoning, understanding ODVs requires models to integrate visual evidence distributed across the surrounding scenes and evolving over time. Therefore, we introduce five spatio-temporal tasks, \textit{i.e.}, \textit{motion perception, event ordering, state tracking, temporal span, and timestamp localization}, to evaluate MLLMs' ability to jointly reason about spatial and temporal information. \textit{Motion perception} and \textit{state tracking} assess the ability to recognize motion and track changes in object states over time. \textit{Event ordering} and \textit{temporal span} evaluate the ability to reconstruct event sequences and estimate their temporal extent. \textit{Timestamp localization} further requires models to associate observations across different viewing directions with specific moments or intervals in the video.
\vspace{-0.5em}
\subsection{Benchmark Construction}
\begin{figure}[t]\centering
\includegraphics[width=1\linewidth]{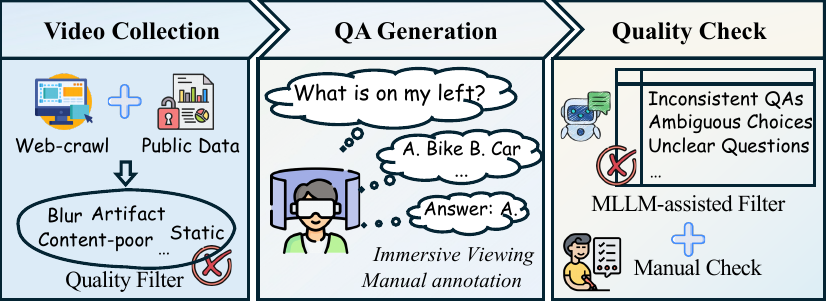}
\vspace{-2em}
\caption{Construction procedure of OmniV-Bench. The videos in our benchmark are carefully selected from web sources and public datasets and undergo rigorous quality filtering to remove low-quality clips. All the question-answer pairs and choices are manually annotated
by human experts through immersive viewing in VR environments. The resulting QA pairs undergo MLLM-assisted filtering and
manual verification to ensure benchmark quality.
}
\label{fig:2}
\vspace{-2em}
\end{figure}
\paragraph{Video Collection.}
The benchmark construction procedure is as illustrated in Figure~\ref{fig:2}. Videos in our benchmark are collected from diverse open sources, including web-crawled content and public datasets~\cite{telili2024odvista,wang2024360dvd,cao2023ntire,li2018bridge,wang2025evoworld} and are carefully selected to ensure both quality and diversity. As presented in Figure~\ref{fig:3}(a), the collected videos cover a wide range of real-world indoor and outdoor environments, including urban streets, public spaces, natural scenes, and indoor activities, thereby providing diverse visual contexts for omnidirectional understanding. To ensure data quality, all videos undergo a data-cleaning pipeline that removes static or content-poor scenes, near-duplicate samples, and videos with noticeable quality degradation, such as severe blurring, extreme exposure, or temporal discontinuities. Long videos are further segmented into shorter clips to retain informative and temporally coherent content while avoiding redundancy.
\vspace{-1em}
\paragraph{Human Annotation.}
Unlike conventional images, where viewers observe a scene from an \textit{external} perspective, omnidirectional media place the viewer at the center of the visual environment and support \textit{immersive} exploration across different viewing directions. To better reflect this viewing paradigm, we formulate questions using first-person and viewer-centric expressions, making the annotations more consistent with how omnidirectional videos are naturally experienced and interacted with. This design also better aligns with practical ODV applications such as embodied perception, spatial navigation, and immersive interaction, \textit{etc.}
All QA pairs in OmniV-Bench, including questions,
answers, and answer choices, are manually annotated
to preserve natural and diverse question formulations. The annotation process involves 30 annotators over one month. Before annotation, all annotators receive dedicated training on the task definitions, annotation criteria, and viewer-centric questioning principles. The annotators then watch the ODVs using VR devices and manually formulate questions and corresponding answers based on the video content. This immersive annotation process allows annotators
to examine the surrounding scenes from different
viewing directions and formulate questions based
on their observations. The resulting QA pairs cover multiple formats, including single-answer multiple-choice, multiple-answer multiple-choice, and yes/no questions.
To ensure annotation quality, all QA pairs undergo multiple rounds of automatic and manual verification. Samples with ambiguous answer choices, unclear question formulations, inconsistent question-answer pairs, or insufficient visual evidence are revised or removed.
\begin{table*}[t]
\centering
\vspace{-1.5em}
\caption{
Benchmark results of MLLMs on OmniV-Bench under the closed-ended
evaluation setting. The tasks are defined as follows:
AR (Attribute Recognition), CNT (Counting), EX (Existence),
OR (ODV Reasoning), EV (Egocentric Viewpoint Understanding),
AV (Allocentric Viewpoint Understanding), MP (Motion Perception),
EO (Event Ordering), ST (State Tracking), TS (Temporal Span),
and TL (Timestamp Localization).
For each task, the best result is indicated in \textbf{bold}
and the second-best result is \underline{underlined}.
}
\vspace{-0.5em}
\label{tab:main_results}


\definecolor{omniOverall}{RGB}{125,103,164}
\definecolor{omniGeneral}{RGB}{74,139,102}
\definecolor{omniSpatial}{RGB}{72,128,181}
\definecolor{omniTemporal}{RGB}{190,155,72}

\definecolor{omniPropBg}{RGB}{216,231,244}
\definecolor{omniOpenBg}{RGB}{228,238,248}
\definecolor{omniSpatialBg}{RGB}{238,245,251}
\definecolor{omniBaselineBg}{RGB}{247,250,253}

\definecolor{omniGroupText}{RGB}{64,91,119}

\renewcommand{\arraystretch}{1.05}
\setlength{\tabcolsep}{5.5pt}
\scriptsize

\resizebox{\textwidth}{!}{%
\begin{tabular}{@{}l*{12}{c}@{}}
\toprule

\multirow{2}{*}{\textbf{Model}}
& \multirow{2}{*}{%
    \textcolor{omniOverall}{\textbf{\textit{Overall}}}%
}
& \multicolumn{4}{c}{%
    \textcolor{omniGeneral}{\textbf{\textit{General}}}%
}
& \multicolumn{2}{c}{%
    \textcolor{omniSpatial}{\textbf{\textit{Spatial}}}%
}
& \multicolumn{5}{c}{%
    \textcolor{omniTemporal}{\textbf{\textit{Spatio-temporal}}}%
} \\
\cmidrule(lr){3-6}
\cmidrule(lr){7-8}
\cmidrule(lr){9-13}

& &{AR} &{CNT} &{EX} &{OR}
  &{EV} &{AV}
  &{MP} &{EO} &{ST}
  &{TS} &{TL} \\

\midrule
\rowcolor{omniPropBg}
\textcolor{omniGroupText}{\bfseries\itshape Proprietary MLLMs}
& & & & & & & & & & & & \\

GPT-4o~\cite{hurst2024gpt}
&54.27 & 71.70 & 49.43 & 75.91 & 75.47 & 54.18 & \underline{46.04} & 55.56 & 33.81 & 55.61 & \textbf{36.62} & 42.68\\

Qwen-3.7-Plus~\cite{qwen3.5}
& \underline{58.11} & \underline{77.52} & 50.76 & 79.17 & \textbf{80.53} & \underline{62.80} & \textbf{47.50} & \underline{66.18} & \underline{42.07} & \underline{63.71} & 23.84 & \textbf{45.16} \\

Gemini-2.5-Flash~\cite{comanici2025gemini}
& 52.73 & 71.77 & 45.21 & 77.29 & 78.08 & 60.75 & 41.25 & 61.09 & 30.34 & 55.61 & 21.19 & 37.50 \\

Gemini-3-Pro~\cite{googledeepmind2026gemini3pro}
& 55.37 & 76.16 & \underline{52.24} & \underline{79.64} & 77.66 & \textbf{63.78} & 43.04 & 60.02 & 22.90 & 60.00 & 29.85 & \underline{43.81} \\

Seed-2.0-Pro~\cite{seed2026seed2}&\textbf{60.25} & \textbf{78.28} & \textbf{59.03} & \textbf{82.07} & \underline{79.89} & 61.71 & \textbf{47.50} & \textbf{72.88} & \textbf{45.52} & \textbf{64.23} & \underline{30.46} & 41.13\\
\midrule
\rowcolor{omniOpenBg}
\textcolor{omniGroupText}{\bfseries\itshape Open-source MLLMs}
& & & & & & & & & & & & \\

LLAVA-Video-7B~\cite{zhang2024llava}&48.85 & 63.84 & 38.85 & 72.40 & 73.91 & 49.05 & 37.75 & 50.55 & 14.48 & 50.13 & \textbf{47.68} & 38.71\\

LLaVA-OV2-8B~\cite{an2026llava}& 52.57 & 71.12 & 43.39 & 76.75 & 76.09 & 56.03 & 36.75 & 51.36 & 31.03 & 49.87 & \underline{46.36} & 39.52 \\

VideoLLaMA2-7B~\cite{damonlpsg2024videollama2}&43.01 & 55.27 & 34.91 & 66.43 & 66.76 & 44.35 & 31.75 & 55.05 & 11.72 & 39.95 & 29.80 & 37.10  \\

VideoLLaMA3-7B~\cite{zhang2025videollama}&51.36 & 68.73 & 44.10 & 73.31 & 74.37 & 54.88 & 39.75 & 59.40 & 13.10 & 55.87 & 37.09 & 44.35\\

GLM-4.1V-9B~\cite{vteam2025glm45vglm41vthinkingversatilemultimodal}
& 38.83 & 42.45 & 27.35 & 68.48 & 38.68 & 42.95 & 30.00 & 51.95 & 4.14 & 49.09 & 37.75 & 34.27\\

Deepseek-VL2-3B~\cite{wu2024deepseek}&40.26 & 56.89 & 30.47 & 66.73 & 63.77 & 45.31 & 35.50 & 51.36 & 6.21 & 39.16 & 19.21 & 28.23 \\

Deepseek-VL2-27B~\cite{wu2024deepseek}&47.46 & 59.61 & 38.24 & 71.14 & 68.21 & 44.48 & 35.25 & 49.23 & 12.41 & 56.66 & \textbf{47.68} & 39.11 \\

Internvideo2.5-8B~\cite{wang2025internvideo2}&48.47 & 62.76 & 41.37 & 69.08 & 71.74 & 52.07 & 41.25 & 52.76 & 13.79 & 59.01 & 35.10 & 34.27 \\

LLaVa-NeXT-7B~\cite{liu2024llavanext}&42.04 & 59.61 & 28.15 & 68.30 & 65.31 & 45.76 & 35.25 & 51.29 & 13.10 & 45.95 & 19.87 & 29.84\\

Qwen2.5-VL-72B~\cite{qwen2.5-VL}
& 53.65 & 70.68 & 45.81 & 73.73 & \underline{77.54} & 58.01 & \textbf{43.50} & 51.22 & 31.03 & 55.09 & 40.40 & 43.15\\

Qwen3-VL-8B~\cite{bai2025qwen3}
&54.28 & 69.82 & 47.43 & 75.30 & 74.64 & 55.78 & 42.25 & 53.43 & 29.66 & 60.05 & 44.37 & 44.35 \\

Qwen3-VL-32B~\cite{bai2025qwen3}
&\underline{56.71}& 74.70 & 51.26 & 77.36 & 76.54 & 61.45 & 41.75 & 59.76 & \underline{35.17} & \underline{65.27} & 41.06 & 39.52 \\

Qwen3.5-4B~\cite{qwen3.5}
&48.47& 68.84 & 42.38 & 74.34 & 71.20 & 55.33 & 39.50 & 59.10 & 11.03 & 53.79 & 20.53 & 37.10 \\

Qwen3.5-27B~\cite{qwen3.5}
&\textbf{59.18} & \underline{74.92} & \underline{51.66} & \underline{78.74} & \textbf{78.53} & \textbf{62.73} & \underline{43.25} & \textbf{68.09} & \textbf{46.21} & \textbf{65.54} & 35.76 & \underline{45.56}\\

Qwen3.5-122B-A10B~\cite{qwen3.5}
&56.63& \textbf{75.24} & \textbf{53.28} & \textbf{78.80} & 77.36 & \underline{62.09} & 41.00 & {60.57} & 33.79 & 63.71 & 33.11 & 43.95\\

InternVL3-8B~\cite{zhu2025internvl3}
&47.31& 65.15 & 36.43 & 69.87 & 68.93 & 47.93 & 36.25 & 54.02 & 19.31 & 48.04 & 35.76 & 38.71 \\

InternVL3-38B~\cite{zhu2025internvl3}
&54.66& 72.53 & 42.89 & 76.21 & 76.36 & 52.78 & 42.00 & 56.74 & 31.03 & 62.40 & 42.38 & \textbf{45.97} \\

InternVL3.5-14B~\cite{wang2025internvl3}
&51.68 & 68.62 & 42.18 & 72.46 & 72.46 & 51.95 & 40.75 & 55.56 & 24.14 & 56.66 & 45.03 & 38.71\\

InternVL3.5-30B-A3B~\cite{wang2025internvl3}
&52.04 & 70.03 & 44.80 & 74.52 & 72.28 & 53.48 & 41.00 & \underline{64.41} & 14.48 & 54.05 & 43.05 & 40.32\\

Deepseek-V4.1-Flash~\cite{xu2026deepseek}
&56.11& 74.05 & 49.04 & 78.02 & 76.63 & 61.14 & 41.25 & 58.44 & 33.79 & 60.31 & 43.05 & 41.53 \\

\midrule
\rowcolor{omniSpatialBg}
\textcolor{omniGroupText}{\bfseries\itshape Spatial-centric MLLMs}
& & & & & & & & & & & & \\

VG-LLM-4B~\cite{zheng2026learningvlm}
&\underline{48.81} & \underline{62.32} & \underline{39.46} & \underline{69.93} & \textbf{73.01} & \underline{49.78} & \textbf{40.25} & 48.49 & 22.07 & 49.87 & \textbf{45.03} & 36.69 \\

VST-3B-SFT~\cite{yang2026visual}
&\textbf{48.96}& \textbf{67.10} & \textbf{40.87} & \textbf{73.73} & \textbf{73.01} & \textbf{53.61} & \underline{39.00} & \textbf{49.96} & 14.48 & 46.21 & \underline{41.06} & \textbf{39.52} \\

Spatial-MLLM~\cite{wu2026spatial}
&25.86& 38.11 & 27.85 & 8.51 & 62.86 & 19.78 & 25.75 & 21.89 & 4.14 & 21.67 & 30.46 & 23.39\\

Cambrian-S-3B~\cite{yang2026cambrian}
&45.94& 54.40 & 35.32 & 65.64 & 70.11 & 47.80 & 35.75 & 48.78 & \underline{22.76} & \underline{52.48} & 34.44 & \underline{37.90}\\

Cambrian-S-7B~\cite{yang2026cambrian}
&46.84& 59.83 & 35.02 & 68.60 & \underline{71.74} & 47.42 & 34.75 & \underline{49.37} & \textbf{25.52} & \textbf{55.09} & 34.44 & 33.47\\

\midrule
\rowcolor{omniBaselineBg}
\multicolumn{13}{@{}l@{}}{%
    \hspace{4pt}%
    \textcolor{omniGroupText}{%
        \bfseries\itshape Baselines%
    }%
} \\

Blind Gemini-2.5-Flash
&40.50 & 41.59 & 41.68 & 52.17 & 41.12 & 43.27 & 37.75 & 54.90 & 15.17 & 45.43 & 39.74 & 32.66\\

Random Choice
&31.94 & 31.03 & 29.15 & 47.62 & 29.41 & 34.07 & 29.68 & 45.07 & 11.19 & 38.15 & 26.66 & 29.33\\

\bottomrule
\end{tabular}%
}
\vspace{-2em}
\end{table*}
\section{Evaluation on OmniV-Bench}
\begin{figure*}[t]\centering
\includegraphics[width=1.0\linewidth]{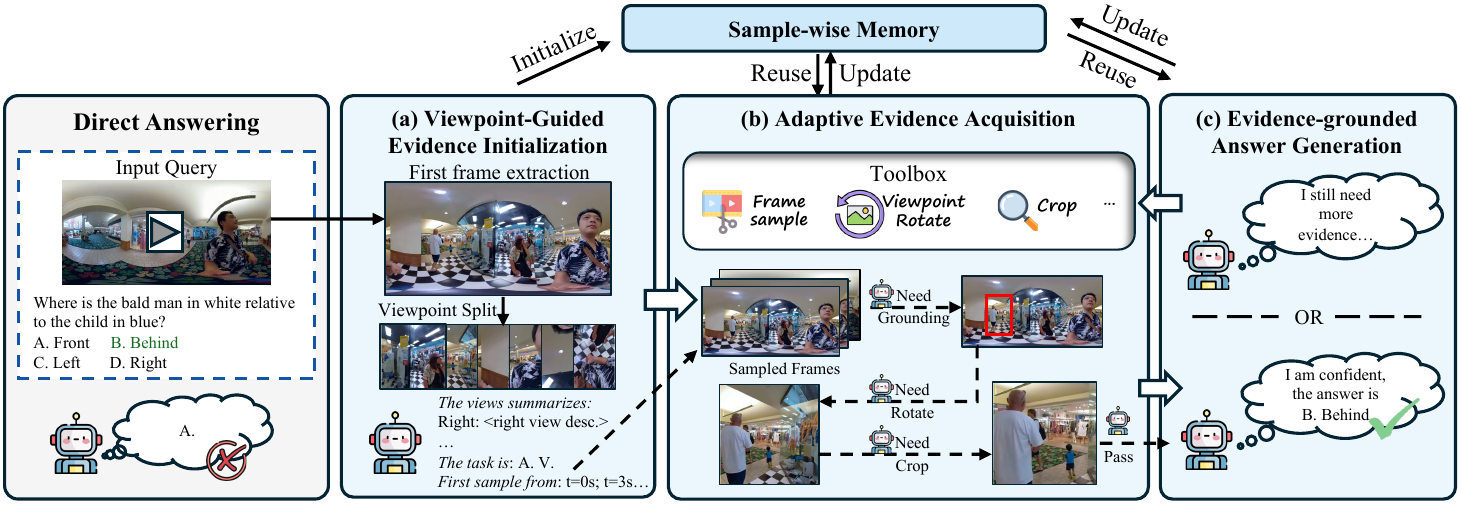}
\vspace{-2em}
\caption{
Overview of OmniV-Harness. \textbf{Left.} direct answering from the input video and query produces an incorrect prediction. \textbf{Right.} OmniV-Harness supports ODV understanding through viewpoint-guided evidence initialization, adaptive evidence acquisition via agentic use of visual tools, and evidence-grounded answer generation. A sample-wise memory maintains the accumulated
visual evidence and tool-use history throughout
the inference process.
}
\label{fig:4}
\vspace{-1.5em}
\end{figure*}

In this section, we conduct extensive benchmark experiment on the proposed OmniV-Bench and provide an in-depth analysis of the experimental results.
\subsection{Evaluation Setup}
We evaluate 30 leading MLLMs with diverse architectures and model scales on OmniV-Bench. The models can be categorized into three groups: (1) proprietary
models, including GPT-4o~\cite{hurst2024gpt}, Gemini~\cite{comanici2025gemini},
etc. (2) open-sourced models, including InternVL series~\cite{zhu2025internvl3, wang2025internvl3}, Qwen-VL series~\cite{bai2025qwen3,qwen2.5-VL,qwen3.5}, LLaVA-NeXT ~\cite{liu2024llavanext}, LLava-OneVision~\cite{an2026llava}, etc. (3) Spatial-centric MLLMs, including VG-LLM~\cite{zheng2026learningvlm}, VST~\cite{yang2026visual}, Spatial-MLLM~\cite{wu2026spatial} and Cambrian-S~\cite{yang2026cambrian}. For a fair comparison, we uniformly sample 50 frames from each ODV and evaluate all models using the same prompt template provided in the Appendix.
Model performance is measured by accuracy on
multiple-choice and yes/no questions.
For event ordering, an answer is considered correct
only when the predicted sequence exactly matches the ground-truth order.
To account for the unequal numbers of questions
across tasks, the overall performance is computed
as the macro-averaged accuracy over all 11 tasks:
\vspace{-0.5em}
\begin{equation}
\mathrm{Acc}_{\mathrm{overall}}
= \frac{1}{|\mathcal{T}|}
\sum_{t \in \mathcal{T}}
\left(
\frac{1}{N_t}
\sum_{i=1}^{N_t}
\mathbb{I}(\hat{y}_{t,i}=y_{t,i})
\right),
\end{equation}
where $\mathcal{T}$ denotes the set of 11 tasks,
$N_t$ is the number of questions in task $t$,
and $\mathbb{I}(\cdot)$ is the indicator function.

\subsection{Main Results}
Benchmark performances of all models across all tasks are reported in Table~\ref{tab:main_results}. We additionally include Blind Gemini-2.5-flash, \textit{i.e., }Gemini-2.5-flash without visual inputs and chance-level accuracy as baselines for reference. Based on these results, several key findings are identified to highlight the limitations of current MLLMs in ODV understanding and provide insights
for future research.
\vspace{-1.5em}
\subsubsection{Overall Performance}

As shown in Table~\ref{tab:main_results}, proprietary MLLMs generally achieve the strongest overall performance, with Seed-2.0-pro ranking first at 60.25\%. Open-source MLLMs also demonstrate competitive performance, with Qwen3.5.27B achieving 59.18\%, approaching leading proprietary models. However, even the best-performing models are still far from satisfactory. These results suggest that:

\begin{findingbox}
ODV understanding remains challenging for MLLMs across different model scales.
\end{findingbox}

Moreover, spatial-centric MLLMs generally underperform leading general-purpose MLLMs on OmniV-Bench. For instance, the best-performing spatial-centric model, VST-3B-SFT, achieves an overall accuracy of 48.96\%, falling behind several general-purpose MLLMs. Despite their explicit focus on spatial understanding, these models do not exhibit a consistent advantage in ODV comprehension.
This observation suggests that:

\begin{findingbox}
ODV understanding presents unique challenges that remain insufficiently addressed by existing spatial reasoning capabilities.
\end{findingbox}

\subsubsection{Task-wise Performance}
As shown in Table~\ref{tab:main_results}, model performance varies considerably across different task categories. For general-level tasks, the dense visual information in immersive scenes poses challenges to basic visual perception, particularly in counting tasks, where models struggle to accurately identify and aggregate objects across the surrounding environment. Performance generally declines on spatial tasks, revealing the limitations of current MLLMs in reasoning about spatial relationships within omnidirectional environments. In particular, allocentric viewpoint understanding requires models to reason from a specified viewpoint, demanding both accurate spatial localization and viewpoint-dependent reasoning across the immersive environment. Compared with egocentric viewpoint understanding, allocentric reasoning exhibits substantially lower accuracy across the evaluated models.

Spatio-temporal tasks introduce even further challenges. While immersive perception of omnidirectional environments already imposes substantial demands on spatial reasoning, incorporating temporal dynamics further challenges the understanding of evolving scenes. Models exhibit particularly poor performance on event ordering, temporal span estimation, and timestamp localization, which require tracking dynamic events and reasoning about their temporal relationships across complex spatial contexts. Performance on several of these tasks remains close to chance level, with only marginal improvements over the blind Gemini baseline in some cases,
revealing substantial limitations in perceiving and reasoning about dynamic events within immersive environments.
Moreover, the considerable performance variation
across models highlights disparities in their
spatio-temporal understanding capabilities.
These observations suggest that:

\begin{findingbox}[findingblue]
Spatial complexity, particularly when coupled
with temporal dynamics, constitutes a major
bottleneck for ODV understanding.
\end{findingbox}
\section{OmniV-Harness}
\label{sec:omni_harness}
In this section, we present \textbf{OmniV-Harness}, an agent harness that enhances MLLMs' comprehension abilities on immersive omnidirectional videos.
\begin{table*}[t]
\centering
\vspace{-1.5em}
\caption{Performance of OmniV-Harness on OmniV-Bench. Improvements over baselines are \textbf{bolded}.}
\vspace{-1em}
\label{tab:omni_harness_results}
\begingroup
\definecolor{omniOverall}{RGB}{125,103,164}
\definecolor{omniGeneral}{RGB}{74,139,102}
\definecolor{omniSpatial}{RGB}{72,128,181}
\definecolor{omniTemporal}{RGB}{190,155,72}
\definecolor{omniHarnessBg}{RGB}{228,238,248}
\definecolor{omniDeltaBg}{RGB}{247,250,253}
\definecolor{omniGroupText}{RGB}{64,91,119}
\scriptsize
\setlength{\tabcolsep}{5.5pt}
\renewcommand{\arraystretch}{1.10}
\resizebox{0.9\textwidth}{!}{%
\begin{tabular}{@{}l*{12}{c}@{}}
\toprule
\textbf{Method} & \textcolor{omniOverall}{\textbf{\textit{Overall}}} & \multicolumn{4}{c}{\textcolor{omniGeneral}{\textbf{\textit{General}}}} & \multicolumn{2}{c}{\textcolor{omniSpatial}{\textbf{\textit{Spatial}}}} & \multicolumn{5}{c}{\textcolor{omniTemporal}{\textbf{\textit{Spatio-temporal}}}} \\
\cmidrule(lr){3-6}\cmidrule(lr){7-8}\cmidrule(l){9-13}
 & & AR & CNT & EX & OR & EV & AV & MP & EO & ST & TS & TL \\
\midrule
\textbf{Qwen3.7-Plus} & 58.11 & 77.52 & 50.76 & 79.17 & 80.53 & 62.80 & 47.50 & 66.18 & 42.07 & 63.71 & 23.84 & 45.16 \\
\rowcolor{omniHarnessBg}
\quad \textcolor{omniGroupText}{\textbf{w/ OmniV-Harness}} & \textbf{62.73} & \textbf{78.94} & \textbf{53.99} & \textbf{79.23} & \textbf{81.34} & \textbf{70.26} & 46.50 & \textbf{70.67} & \textbf{44.14} & 60.57 & \textbf{50.33} & \textbf{54.03} \\
\rowcolor{omniDeltaBg}
\quad $\Delta$ (pp) & +4.62 & +1.41 & +3.23 & +0.06 & +0.82 & +7.47 & -1.00 & +4.50 & +2.07 & -3.13 & +26.49 & +8.87 \\
\midrule
\textbf{Gemini-2.5-Flash} & 52.71 & 71.77 & 45.21 & 77.29 & 78.08 & 60.75 & 41.00 & 61.09 & 30.34 & 55.61 & 21.19 & 37.50\\
\rowcolor{omniHarnessBg}
\quad \textcolor{omniGroupText}{\textbf{w/ OmniV-Harness}} & \textbf{59.55} & \textbf{77.85} & \textbf{51.46} & \textbf{80.31} & \textbf{80.98} & \textbf{66.94} & \textbf{43.50} & \textbf{67.13} & \textbf{31.03} & \textbf{58.49} & \textbf{45.70} & \textbf{51.61} \\
\rowcolor{omniDeltaBg}
\quad $\Delta$ (pp) & +6.84 & +6.08 & +6.26 & +3.02 & +2.90 & +6.19 & +2.50 & +6.04 & +0.69 & +2.87 & +24.50 & +14.11 \\
\midrule
\textbf{Qwen3.5-4B} & 48.47 & 68.84 & 42.38 & 74.34 & 71.20 & 55.33 & 39.50 & 59.10 & 11.03 & 53.79 & 20.53 & 37.10 \\
\rowcolor{omniHarnessBg}
\quad \textcolor{omniGroupText}{\textbf{w/ OmniV-Harness}} & \textbf{56.33} & \textbf{76.44} & \textbf{49.04} & \textbf{80.13} & \textbf{80.43} & \textbf{63.05} & 36.75 & \textbf{64.11} & \textbf{21.38} & \textbf{55.61} & \textbf{46.36} & \textbf{46.37} \\
\rowcolor{omniDeltaBg}
\quad $\Delta$ (pp) & +7.87 & +7.60 & +6.66 & +5.80 & +9.24 & +7.72 & -2.75 & +5.01 & +10.34 & +1.83 & +25.83 & +9.27 \\
\midrule
\textbf{InternVL3-8B} & 47.31 & 65.15 & 36.43 & 69.87 & 68.93 & 47.93 & 36.25 & 54.02 & 19.31 & 48.04 & 35.76 & 38.71 \\
\rowcolor{omniHarnessBg}
\quad \textcolor{omniGroupText}{\textbf{w/ OmniV-Harness}} & \textbf{54.68} & \textbf{73.94} & \textbf{42.68} & \textbf{74.40} & \textbf{77.17} & \textbf{56.41} & \textbf{39.25} & \textbf{61.97} & \textbf{27.59} & \textbf{54.31} & \textbf{47.02} & \textbf{46.77} \\
\rowcolor{omniDeltaBg}
\quad $\Delta$ (pp) & +7.38 & +8.79 & +6.26 & +4.53 & +8.24 & +8.49 & +3.00 & +7.96 & +8.28 & +6.27 & +11.26 & +8.06 \\
\bottomrule
\end{tabular}%
}
\endgroup
\vspace{-1.5em}
\end{table*}

\subsection{Framework Overview}
Compared with conventional video understanding, omnidirectional videos introduce additional
challenges, as task-relevant evidence may be distributed
across different viewing directions and timestamps,
while substantial visual redundancy in immersive
spatio-temporal scenes makes it difficult to identify
informative observations.

To address these challenges, OmniV-Harness performs viewpoint-guided evidence initialization to establish directional context and adaptive evidence aquisition to select task-relevant evidence, as presented in Figure~\ref{fig:4}. The framework first establishes directional context from omnidirectional observations to facilitate subsequent spatial reasoning. Guided by the input query and accumulated evidence, MLLMs then adaptively explore relevant temporal timestamps, viewing directions, and cropped regions through iterative tool use. Finally, the framework integrates the acquired evidence to generate predictions, with further visual inspection when the available evidence is insufficient or conflict. Throughout this process, a sample-wise memory maintains the collected observations and intermediate reasoning states to support iterative evidence acquisition and answer refinement.
\subsubsection{Viewpoint-Guided Evidence Initialization}
Directly interpreting the equirectangular representation of ODVs makes the immersive
directional information difficult to identify. Therefore,
we adopt viewpoint-guided mechanism for evidence initialization. Specifically, the first
ODV frame is projected into six perspective views (front,
back, left, right, up, and down), each with a $90^\circ$
field of view. In a single MLLM call,
the model generates a brief description of each view,
predicts the task type, and identifies observation
requirements, including candidate sampling timestamps.
These model-suggested timestamps are combined with
timestamps uniformly sampled at a low sampling rate
and temporal references from the query to construct
the initial observations. The resulting visual evidence
and directional context are maintained in the
sample-wise memory to guide subsequent evidence
acquisition.

\subsubsection{Adaptive Evidence Acquisition}
Given the initialized visual evidence and directional
context, OmniV-Harness enables the MLLM to determine
whether and where additional visual evidences are needed.
Rather than relying on a fixed set of densely sampled
video frames, the model selectively explores the
video according to the input query and current evidence.

To support adaptive visual exploration, we provide
three complementary perception tools.
(1) Temporal sampling allows the model to inspect
additional timestamps to capture missing events or
refine temporal boundaries.
(2) Viewpoint rotation enables the model to
explore different viewing directions and fields of
view, facilitating spatial reasoning in immersive
environments.
(3) Region cropping further supports fine-grained
inspection of task-relevant objects and local regions.
Together, these tools enable selective evidence
acquisition across the spatial and temporal dimensions
of ODVs.
At each iteration, the MLLM issues structured tool
requests, which are validated and executed within
predefined budgets. The acquired observations,
together with their timestamps and directional context,
are incorporated into the sample-wise memory.
The updated evidence then guides subsequent tool use,
allowing the model to progressively focus on informative
spatio-temporal observations rather than following
a fixed sequence of visual operations.
\vspace{-1.5em}
\subsubsection{Evidence-Grounded Answer Generation}
Based on the acquired visual evidence, OmniV-Harness
enables the MLLM to generate a candidate answer by
integrating the observations and intermediate reasoning
states maintained in the sample-wise memory.
The model then assesses whether the available evidence
sufficiently supports its prediction, with task-specific
consistency checks applied when necessary to resolve
ambiguities in spatial relationships and temporal events.

When additional evidence is needed, the model returns
to the adaptive evidence acquisition stage to gather
further observations and refine its prediction based
on the updated evidence.
This iterative review-and-refinement process continues
until the model determines that sufficient evidence
is available or the predefined interaction budget
is reached.
\vspace{-0.5em}
\section{OmniV-Harness Performance}
\subsection{Performance Improvement on OmniV-Bench}
We evaluate OmniV-Harness on four representative MLLMs,
including two proprietary models, Qwen3.7-Plus and
Gemini-2.5-Flash, and two open-source models,
Qwen3.5-4B and InternVL3-8B.
As shown in Table~\ref{tab:omni_harness_results},
OmniV-Harness consistently improves overall ODV
understanding performance for both proprietary and
open-source models of varying scales.
The improvements are particularly pronounced on
spatio-temporal tasks, where accuracy on
\textit{temporal span} more than doubles for three
models, while \textit{timestamp localization} alsox
achieves substantial gains.
The results demonstrate the effectiveness of
OmniV-Harness in enhancing MLLMs' understanding of
immersive spatio-temporal scenes.

\subsection{Performance Comparison with Other Harnesses}
\vspace{-0.5em}
\begin{figure}[t]\centering
\includegraphics[width=1\linewidth]{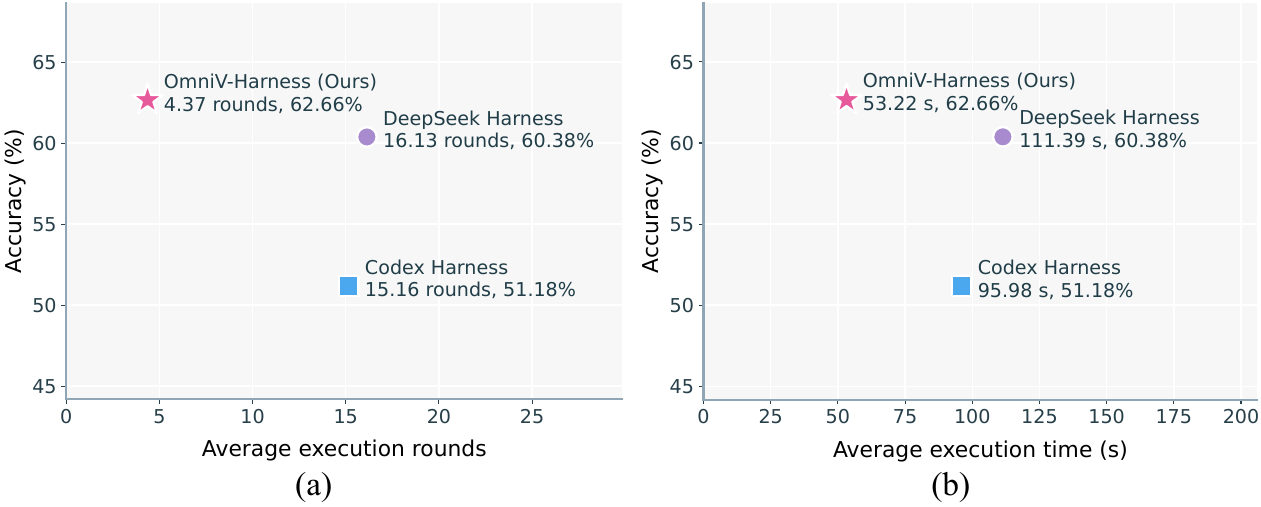}
\vspace{-2em}
\caption{Comparison between Codex, Deepseek-Harness and OmniV-Harness (ours) on OmniV-Bench.}
\label{fig:5}
\vspace{-1.5em}
\end{figure}
To comprehensively evaluate the effectiveness and efficiency of different harness frameworks for ODV understanding, we randomly sample 100 questions from each task for comparison. For a fair comparison, all harnesses employ Qwen3.7-Plus as the execution model with identical model configurations. Notably, we impose no additional restrictions on the permissions or tool usage of DeepSeek Harness or Codex Harness, allowing them to operate with their full native capabilities.

As shown in Fig.~\ref{fig:5}, OmniV-Harness achieves
the highest overall accuracy on OmniV-Bench, outperforming both DeepSeek Harness and Codex Harness. Moreover, OmniV-Harness requires only 4.37 execution rounds and 53.22 seconds per question on average, representing reductions of 72.9\% in execution rounds and 52.2\% in inference time compared with DeepSeek Harness. These results demonstrate that OmniV-Harness achieves superior performance with substantially lower execution overhead, highlighting the advantages of our ODV-oriented harness design for immersive spatio-temporal understanding.
\vspace{-0.5em}
\subsection{Ablation Study}
To further evaluate the effectiveness of OmniV-Harness, comprehensive ablation studies are conducted to examine the contributions of its core framework components and visual tools.
\subsubsection{Module-wise Ablation Study}

\begin{table}[t]
\centering
\caption{
Module-wise ablation study of OmniV-Harness on OmniV-Bench.
VGI, AEA, and AC denote Viewpoint-Guided Evidence
Initialization, Adaptive Evidence Acquisition,
and Answer Check and Refinement Mechanism in Evidence-grounded Answer Generation, respectively.
}
\vspace{-1em}
\label{tab:omni_harness_ablation}

\renewcommand{\arraystretch}{1.10}
\setlength{\tabcolsep}{5pt}
\setlength{\arrayrulewidth}{0.35pt}
\scriptsize

\resizebox{\columnwidth}{!}{%
\begin{tabular}{l|ccc|cccc}
\specialrule{1.0pt}{0pt}{0pt}

\multirow{2}{*}{\textbf{Model}}
& \multicolumn{3}{c|}{\textbf{Module}}
& \multicolumn{4}{c}{\textbf{Performance}} \\
\cmidrule(lr){2-4}
\cmidrule(lr){5-8}
& VGI & AEA & AC
& Overall & General & Spatial & Spatio-temp. \\
\hline

Qwen3.7-Plus
& & &
&58.11 &72.00&55.15&48.19 \\

& \ding{51} & &
&61.12&71.09&56.46&55.00\\

& \ding{51} & \ding{51} &
&62.24&73.37&57.28&55.33\\

&& \ding{51}  & \ding{51}
&61.52&71.84&55.07& 55.83 \\

\rowcolor{gray!12}
& \ding{51} & \ding{51} & \ding{51}
&\textbf{62.73}&\textbf{73.38}&\textbf{58.38}&\textbf{55.95}\\

\hline

Qwen3.5-4B
& & &
&48.47 & 64.19 & 47.42 & 36.31\\

& \ding{51} & &
&53.21 & 64.66 & 49.50 & 45.54 \\

& \ding{51} & \ding{51} &
&53.61&65.88&46.66&46.59\\

&&  \ding{51} & \ding{51}
&53.75&67.03&46.86&45.88\\

\rowcolor{gray!12}
& \ding{51} & \ding{51} & \ding{51}
&\textbf{56.33} & \textbf{71.51} & \textbf{49.90} & \textbf{46.77} \\

\specialrule{1.0pt}{0pt}{0pt}
\end{tabular}
}
\vspace{-1em}
\end{table}

We conduct module-wise ablation studies on Qwen3.7-Plus and Qwen3.5-4B to evaluate the contributions of individual components in OmniV-Harness.
As reported in Table~\ref{tab:omni_harness_ablation}, comparing the first two configurations demonstrates the effectiveness of our viewpoint-guided evidence initialization tailored for omnidirectional understanding, particularly on spatio-temporal tasks. Further incorporating adaptive evidence acquisition enables MLLMs to selectively explore richer task-relevant visual evidence, leading to improved
performance. Moreover, the answer checking and refinement
mechanism in evidence-grounded answer generation provides additional gains by assessing
evidence sufficiency and revisiting relevant
observations when necessary.
\subsubsection{Tool-wise Ablation Study}

\begin{table}[t]
\centering
\caption{
Tool-wise ablation study of OmniV-Harness on OmniV-Bench
using Qwen3.5-4B.
FS, VR, and RC denote Frame Sampling,
Viewpoint Rotation, and Region Cropping, respectively.
}
\vspace{-0.8em}
\label{tab:omni_harness_tool_ablation}

\renewcommand{\arraystretch}{1.10}
\setlength{\tabcolsep}{5pt}
\setlength{\arrayrulewidth}{0.35pt}
\scriptsize

\resizebox{0.9\columnwidth}{!}{%
\begin{tabular}{ccc|cccc}

\specialrule{1.0pt}{0pt}{0pt}

\multicolumn{3}{c|}{\textbf{Visual Tool}}
& \multicolumn{4}{c}{\textbf{Performance}} \\

\cmidrule(lr){1-3}
\cmidrule(lr){4-7}

FS & VR & RC
& Overall & General & Spatial & Spatio-temp. \\

\hline

& \ding{51} & \ding{51}
&53.04&69.47&46.42&42.56\\

\ding{51} & & \ding{51}
&52.55&69.01&45.79&42.07\\

\ding{51} & \ding{51} &
&52.43&68.17&47.31&41.88\\

\rowcolor{gray!12}
\ding{51} & \ding{51} & \ding{51}
&\textbf{56.33} & \textbf{71.51}& \textbf{49.90} & \textbf{46.77} \\

\specialrule{1.0pt}{0pt}{0pt}

\end{tabular}%
}
\vspace{-2em}
\end{table}

Tool-wise ablation studies are also conducted
using Qwen3.5-4B to evaluate the contributions of
individual visual tools within the adaptive evidence
acquisition module.
As shown in Table~\ref{tab:omni_harness_tool_ablation},
removing any visual tool leads to performance
degradation, demonstrating their complementary
roles in supporting omnidirectional video
understanding.
\vspace{-0.5em}
\section{Conclusion}
\vspace{-0.5em}
In this work, we first introduce OmniV-Bench, a comprehensive benchmark for evaluating MLLMs' understanding capabilities of immersive omnidirectional videos. OmniV-Bench comprises 4,123 high-quality ODVs along with 8,923 human-annotated question-answer pairs spanning 11 fine-grained tasks across general, spatial, and spatio-temporal categories. 30 leading MLLMs are comprehensively evaluated, revealing substantial limitations in their ability to comprehend immersive spatio-temporal environments. To address these limitations, we further propose OmniV-Harness, an agent harness that enhances MLLMs' ODV understanding through viewpoint-guided evidence initialization and adaptive evidence aquisition. Together, OmniV-Bench and OmniV-Harness highlight the challenges of immersive spatio-temporal understanding and offer a promising direction for advancing MLLMs' capabilities in omnidirectional environments.
{
    \small
    \bibliographystyle{ieeenat_fullname}
    \bibliography{main}
}


\end{document}